\documentclass{article}

\usepackage[utf8]{inputenc}
\usepackage{amsmath}
\usepackage{amsfonts}
\usepackage{enumerate}
\usepackage{amssymb}
\usepackage{amsthm}
\usepackage{verbatim}
\usepackage{hyperref}
\usepackage{etoolbox}
\usepackage{bm}
\usepackage[capitalize,noabbrev]{cleveref}
\usepackage{graphicx}
\usepackage{comment}
\usepackage{color}
\usepackage{parskip}
\usepackage{mathtools}
\usepackage{multirow}
\usepackage{microtype}
\usepackage{colortbl}
\usepackage{booktabs} %
\usepackage{dcolumn}  
\usepackage{multirow} 
\usepackage{framed}
\usepackage{varwidth}
\usepackage{xspace}
\usepackage{nicefrac}
\usepackage{color}
\usepackage{amssymb}
\usepackage{pifont}
\usepackage{relsize}
\usepackage[operators, ff, adversary]{cryptocode}
\usepackage{extarrows}
\usepackage{subcaption}
\usepackage{pgfplots}
\pgfplotsset{compat=1.18}

\usepackage[normalem]{ulem}
\usepackage{peanutgallery}
\usepackage{outlines}
\usepackage{tcolorbox}

\usepackage{listings}
\usepackage{ulem}
\theoremstyle{plain}

\newcommand{\ignore}[1]{{}}

\usepackage{tikz}
\tikzset{
every picture/.style={thick} 
}
\usetikzlibrary{positioning,matrix,calc,arrows,shapes,fit,decorations,angles,quotes}

\def\hn{\usefont{OT1}{phv}{mc}{n}\selectfont}

\newcommand{\doubleplus}{+\kern-1.3ex+\kern0.8ex}

\def\compactify{\itemsep0in \topsep2pt \parsep=0.00in \partopsep=0pt
\leftmargin2em}
\let\latexusecounter=\usecounter

  {\begin{list}{\labelitemi}{\itemsep0in \topsep2pt \parsep0.00in
  \partopsep=0pt \leftmargin\parindent}}%
  {\end{list}}

  {\begin{list}{\labelitemi}{\itemsep6pt \topsep6pt \parsep0.00in
  \partopsep=0pt \leftmargin\parindent}}%
  {\end{list}}

  {\begin{list}{\labelitemi}{\itemsep3pt \topsep3pt \parsep0.00in
  \partopsep=0pt \leftmargin\parindent}}%
  {\end{list}}
  {\begin{list}{\labelitemi}{\itemsep0in \topsep2pt \parsep0.00in
  \partopsep=0pt \leftmargin\parindent}}%
  {\end{list}}

\usepackage{algorithmic}
\usepackage{algorithm}

\usepackage{inconsolata}
\definecolor{darkspringgreen}{rgb}{0.09, 0.45, 0.27}
\newcommand{\raisedasterisk}{\raisebox{0.25ex}{* }}
\tcbset{
    mycodebox/.style={
        colback=gray!10, %
        colframe=gray!50, %
        boxrule=0.5mm, %
        arc=1mm, %
        outer arc=1mm,
        boxsep=1mm, %
        left=1mm, %
        right=1mm, %
    }
}

\makeatletter
\def\blfootnote{\gdef\@thefnmark{}\@footnotetext}
\makeatother

\newcommand{\ieeefp}{IEEE-754\xspace}

\newcommand{\trainer}{trainer\xspace}
\newcommand{\trainers}{trainers\xspace}

\newcommand{\tth}{\mathtt{h}\xspace}

\newcommand{\repops}{{RepOps}\xspace}

\newcommand{\cuda}{\textsf{CUDA}\xspace}
\newcommand{\cudnn}{\textsf{cuDNN}\xspace}

\newcommand{\pytorch}{\text{PyTorch}\xspace}

\newcommand{\autograd}{{autograd}\xspace}
\newcommand{\onnx}{{ONNX}\xspace}

\newcommand{\cgraphnode}{\texttt{AugmentedCGNode}\xspace}

\newcommand{\distilbert}{\text{DistilBERT}\xspace}
\newcommand{\llama}{\text{Llama}\xspace}

\newcommand{\server}{\text{\trainer}\xspace}
\newcommand{\servers}{\text{\trainers}\xspace}

\newcommand{\node}{\mathsf{node}\xspace}

\definecolor{indianred}{rgb}{0.8, 0.36, 0.36}
\definecolor{indiagreen}{rgb}{0.07, 0.53, 0.03}
\definecolor{jade}{rgb}{0.0, 0.66, 0.42}
\definecolor{azure}{rgb}{0.0, 0.5, 1.0}

\usepackage[accepted]{icml2025}

\theoremstyle{plain}

\theoremstyle{definition}

\theoremstyle{remark}

\usepackage[textsize=tiny]{todonotes}

\icmltitlerunning{Verde: Verification via Refereed Delegation for Machine Learning Programs}

\begin{document}

\twocolumn[
\icmltitle{Verde: Verification via Refereed Delegation for Machine Learning Programs}

\icmlsetsymbol{equal}{*}

\begin{icmlauthorlist}
\icmlauthor{Arasu Arun}{gensyn,nyu}
\icmlauthor{Adam St. Arnaud}{gensyn}
\icmlauthor{Alexey Titov}{gensyn}
\icmlauthor{Brian Wilcox}{gensyn}
\icmlauthor{Viktor Kolobari\'{c}}{gensyn}
\icmlauthor{Marc Brinkmann}{gensyn}
\icmlauthor{Oguzhan Ersoy}{gensyn}
\icmlauthor{Ben Fielding}{gensyn}
\icmlauthor{Joseph Bonneau}{nyu,a16z}
\end{icmlauthorlist}

\icmlaffiliation{gensyn}{Gensyn AI}
\icmlaffiliation{nyu}{New York University}
\icmlaffiliation{a16z}{a16z crypto research}

\icmlcorrespondingauthor{Arasu Arun}{arasu@nyu.edu}

\vspace{0.25cm}
\centerline{$^1$Gensyn AI \quad $^2$New York University \quad $^3$a16z crypto research}
\vspace{0.5cm}

\icmlkeywords{Machine Learning, ICML}

]

\renewcommand\thefootnote{}
\footnotetext{Correspondence to $<$arasu@gensyn.ai$>$}
\renewcommand\thefootnote{\arabic{footnote}}

\begin{abstract}
Machine learning programs, such as those performing inference, fine-tuning, and training
of LLMs, are commonly delegated to untrusted compute providers. 
To provide correctness guarantees for the client, we propose adapting the cryptographic notion of refereed delegation
to the machine learning setting.
This approach enables a computationally limited client to delegate 
a program to multiple untrusted compute providers,
with a guarantee of obtaining the correct result if at least one of them is honest.
Refereed delegation of ML programs poses two technical hurdles:
(1) an arbitration protocol to resolve disputes when compute providers disagree on the output,
and (2) the ability to bitwise reproduce ML programs across different hardware setups,
For (1), we design Verde, a dispute arbitration protocol
that efficiently handles the large scale and graph-based computational model
of modern ML programs.
For (2), we build RepOps (Reproducible Operators),
a library that eliminates hardware ``non-determinism'' by
controlling the order of floating point operations performed
on all hardware.
Our implementation shows that refereed delegation achieves 
both strong guarantees for clients and practical overheads for compute providers. 

\end{abstract}

\section{Introduction}

Machine learning can be expensive, in both computation and memory requirements.
As a result, it is common to delegate many tasks involving neural networks---such 
as training, fine-tuning, or inference---to external compute providers with significant capacity 
and specialized hardware.
In practice this delegation is typically done with no guarantees of correctness.
A dishonest or compromised server might return incorrect outputs.
For example, a lazy server might not train an LLM for the duration claimed and return an approximate result.
An outright malicious \server could perform data poisoning attacks or insert a backdoor into the trained model.

How can clients be assured that the results they receive is honest? 
Cryptographic proof systems offer the strongest guarantee 
solution to this problem, enabling any untrusted party to produce a short and easy-to-check mathematical proof that they ran an arbitrary program correctly. 
While these proof systems are generic and can be applied directly to ML tasks, they are inefficient.
Even using state-of-the-art proof techniques, computing the proof is at least 4 orders of magnitude more expensive than running the program itself. 
For example, for a 125 million parameter neural network, a proof of inference for a single forward pass takes over a minute~\cite{sun24zkllm}, and for a training step takes about 15 minutes~\cite{abbaszadeh24zero}. 
This limits the practical application of these techniques to only very small, extremely valuable ML tasks.

Heuristic, ML-specific techniques, such as Proof-of-Learning~\cite{jia21proofoflearning} 
or Proof-of-Training-Data~\cite{choi24tools}, 
trade off the formal guarantees of cryptographic proofs for efficiency.
In both methods, model trainers log and share intermediate checkpoints with the verifiers. 
Under the assumption that valid executions tend to result in small weight updates, 
verifiers in Proof-of-Training re-execute the training segments 
representing the largest magnitude changes in weights. 
In Proof-of-Training-Data, verifiers query checkpoints to check if the models ``memorized'' 
(i.e., overfit to) the data they were just trained on.
These techniques are narrow in scope 
and lack rigorous security guarantees. 
Indeed, many practical attacks have been demonstrated~\cite{fang23proof}. 
As a result these methods are perhaps only applicable to low-value tasks.

Thus, we're still motivated to look for a method to provide concrete guarantees while imposing practical overheads on compute providers. 
A simple approach is for a client to delegate the same task to multiple compute providers.
Naively, the client could take a majority vote and ensure correctness assuming most of the servers are honest.
In fact, the client can do better and ensure correctness if \emph{any single compute provider} is honest.
The key idea is efficient \emph{dispute resolution}: if two servers produce different output, the client needs to determine which is incorrect.
A naive approach is to ask a trusted party, the ``referee'', to re-run the computation to determine the right output.
However, if we had such a trusted referee with sufficient capacity, the client could have simply asked them to run the computation in the first place.
Therefore we are most interested in the case where the referee is highly limited computationally, for example the client requesting the work themselves also acts as the referee.

Canetti et al.~\cite{canetti13refereed} shows an approach to significantly reduce the work done by the referee based on a simple observation: if two parties disagree on the output of a program, there must be some \emph{point of divergence} in their computation path. 
That is, there is some program step in which they start with the same state (for a CPU program, for example, this state would comprise all contents of registers and memory) but end up with different states. 
For short computations, each compute provider might send their full sequence of program states (or \emph{computation trace}) to the referee.
For longer computations, Canetti et al.~\cite{canetti13refereed} proposed using interaction
between the referee and the compute providers and succinct \emph{commitments} to program states
to efficiently find the point of divergence via binary search, without sending the full computation trace to the referee.
Once it is found, the referee need only re-execute a single computation step at the point of divergence, which by definition at least one of the compute providers must have executed incorrectly.
Hence, the referee can prove that at least one party is incorrect.

In this work, we apply the refereed delegation model to ML programs. This requires overcoming two main challenges:

\textbf{Challenge 1: Adapting dispute resolution to ML programs.} 
Existing refereed delegation protocols are designed for CPU programs and do not efficiently translate to the scale of modern neural networks.
Program states, consisting of neural network parameters, are many gigabytes in size, and program steps involve highly parallelized code (often run on GPUs or similar hardware).
We must also enumerate all possible deviations from honest behavior, and design commitments to intermediate program checkpoints that allow only an honest server to prove that their computation was done correctly.
We propose Verde, a dispute resolution protocol tailored to modern ML programs, such as the training of LLMs. 
We treat neural network programs as an abstract state machine defined by learnable parameters and a computational graph. 
We narrow down the dispute in two phases: the first narrows it down to a single training or inference step, and the second narrows it down to a single operator in the graph.
The referee's only needs to compute a single operator in the computational graph, which can be performed with two orders of magnitude less compute resources than it takes to run the model itself. 
The \servers are only required to store and hash intermediate training checkpoints on top of re-executing small segments of training, thus incurring low overheads.

\paragraph{Challenge 2: Ensuring bitwise reproducibility across hardware setups.}
The refereed delegation model assumes that honest servers will always compute the same result for the same program, but this may not be true if they are using different hardware.
In particular, hardware may provide different numerical results because floating point operations are not guaranteed to be associative, even when following the \ieeefp standard: $(a+b)+c$ and $a+(b+c)$ can be different when the operands are floating-point values.
Given the same parallelized program,
owing to differences in architecture, distinct hardware implementations may (and often do) execute $a+b+c$
in different orders, ending up with different results.
To overcome this, we design \repops (Reproducible Operators), a library that implements bitwise reproducible versions of popular ML operators. 
It works by ensuring that floating point operations are performed in the same order, eliminating hardware non-determinism.
\repops currently provides CUDA implementations of all the operators needed to run \distilbert and the \llama family of models. 
We find that the overhead of \repops is moderate.
In fact, when comparing against \pytorch using cuDNN~\cite{chetlur14cuDNN} implementations of operators,
\repops matrix multiplications (of dimensions larger than $2^{10}$) incur less than 30\% overhead 
when tested on various GPUs. 
Similarly, 
both inference and training on \llama-3.1-1B incur around 60\% overhead \repops 
on an A100 GPU (with 40 GB VRAM).

By definition, as refereed delegation requires multiple servers executing the same program, 
it incurs at least a factor 2 total overhead relative to the program itself. 
However, even with the use of \repops, 
this refereed delegation can be achieved with totally less than an order of magnitude overhead. 
This is dramatically more efficient than cryptographic proofs (4 orders of magnitude).
Our approach could be used for opportunistic auditing of delegated computation 
with a client serving as a referee, 
or as a building block in decentralized training applications 
with an on-chain smart contract serving as a referee.

\section{Verde: Dispute resolution for neural networks}
\label{sec:game}

Consider a client delegating the same ML program (e.g. training, inference) to $k>1$ compute providers, whom we'll generically call `\servers'.
Each \server responds with a claimed output (or a binding commitment to an output).
If all \servers return the same output, then there is no need for dispute resolution.
If the \servers disagree on the output, they'll engage in a dispute resolution protocol with a neutral referee (which could be the client themselves).
For simplicity, we assume all $k$ \servers claims a different output. Any \servers claiming the same output can be merged into one.
At the end of the protocol, the referee selects one of the $k$ outputs to accept. 

The security guarantee of dispute resolution is as follows:
If there is an honest \server, their output is guaranteed to be the one accepted.
The referee will identify $k-1$ dishonest \servers, proving that they reported incorrect output.
An important limitation is that if \emph{all} \servers are dishonest, the protocol will still identify $k-1$ dishonest providers, but the referee will also an incorrect output and not find proof of incorrectness for one dishonest provider.

{\bf Program setup:} 
The client specifies a neural network with initial weights, training data, and training metadata 
  like the optimizer and global batch size. 
The model is a topologically-sorted computational graph, 
  consisting of operators (using standardized formats like \onnx\cite{onnx2017}). 
A ``training step'' comprises a forward pass, backward pass, parameter updates 
  and an optimizer state update. 

For simplicity of presentation, we consider only the case of $k=2$ \servers.\footnote{Extending to $k>2$ \servers can be achieved trivially by repeating the 2-\server case iteratively or in a hierarchical tournament. More efficient combination is possible by combining some steps.}
We assume bitwise reproducibility between the referee and all \servers, ensuring that honest execution 
always produces the same output (Section~\ref{sec:repops} explains how \repops enables this 
on different hardware setups).
The referee has access to the computational graph defining the model, the initial parameters, and the training data. 
However, they are assumed to be computationally limited, and unable to train the model themselves. 

During dispute resolution, the referee interacts with the \servers, 
analyzing commitments to their intermediate computation states to narrow down the 
dispute to a smaller segment of the program.
In adapting this to neural network training, we design 
a protocol that proceeds in two phases.
\begin{enumerate}
  \item {\bf Phase 1} identifies the exact training step 
    that the \trainers first diverged at.
  \item {\bf Phase 2} identifies the first operator in the computational graph
    defining the model where they diverge. 
\end{enumerate}

Sections \ref{sec:bisection-phase-1} and \ref{sec:bisection-phase-2} 
describe the two phases, respectively.
While we focus on training, our techniques immediately extend 
to inference. 

\subsection{Phase 1: Identifying the diverging training step.}
\label{sec:bisection-phase-1}

A training program specifies the neural network as a computational graph 
with a training dataset and fixed training metadata. 
We start by abstracting the training procedure as a state machine, where the 
``state'' is the values of the learnable parameters and optimizer state 
and the ``transition function'' is each training step 
performing a forward pass, backward pass, and optimizer state update. 
At each training step, we derive a ``checkpoint'' that includes the state and 
can be \emph{committed} using a cryptographically binding commitment scheme. 
We define a precise checkpoint and commitment format later, 
but it suffices for now to assume that the checkpoint consists of just the aforementioned state 
and is committed to using a standard collision-resistant hash function like SHA-256. 

The training program transcript can thus be viewed as a sequence of checkpoints 
$C_0 \rightarrow C_1 \rightarrow \ldots \rightarrow C_n$. 
If two trainers start with $C_0$ 
but produce different outputs after $n$ steps, $C_n$ and $C_n'$, 
then there must be a training step $i$ where they first ``diverged'':
that is, they claim different transitions 
$C_i \rightarrow C_{i+1}$ and $C_i \rightarrow C_{i+1}'$, 
respectively, at step $i$. 

{\bf A first attempt.} 
The naive way for the referee to identify this diverging step is to 
receive from each \trainer the 
hashes of all their checkpoints, and then find the diverging checkpoint with a linear search.
The prohibitive issue here is that hashing 
the model state after every single step can be very expensive for large models.
For example, hashing the weights and Adam optimizer state~\cite{kingma14adam}
(the optimizer state is double the size of the weights alone)
in FP32 precision for 
\distilbert (66 million parameters) takes under a second, 
for \llama-1B takes around 2.5 seconds, and for \llama-8B around 15 seconds, 
on a CPU with 128 GB RAM and the Apple M3 chip.

\newcommand{\starthash}{\mathtt{h_{start}}\xspace}
\newcommand{\hend}{\mathtt{{h}_{end}}\xspace}
\newcommand{\disphash}{\mathtt{{h}_{end}}\xspace}
\newcommand{\disphashes}{\mathtt{\overline{h}_{end}}\xspace}

\begin{algorithm}
  \begin{algorithmic}[1]
\STATE {\bf Participants}: Trainers $T_0, T_1$, Referee $R$
\STATE {\bf Inputs}: Starting checkpoint $C_0$, with hash $\starthash$. 
\STATE {\bf Parameter}: Checkpoint counts $k_0, k_1, k_2, \ldots$
\STATE Trainers send hash of output checkpoint to referee.
\STATE \textbullet\ $T_0\rightarrow R$: $\hend_1$
\STATE \textbullet\ $T_1\rightarrow R$: $\hend_2$
\STATE $R$: If $\hend_1 == \hend_2$, then output no dispute.
\STATE Let $\disphashes = [\hend_1, \hend_2]$
\FOR{$\ell$ = 0, 1, \ldots}
  \STATE Trainers send $k_\ell$ checkpoint hashes to $R$: 
  \STATE \textbullet\ $T_0\rightarrow R$: $\starthash, \tth_1, \ldots, \disphashes[0]$
  \STATE \textbullet\ $T_1\rightarrow R$: $\starthash, \tth_1', \ldots, \disphashes[1]$
  \STATE Referee identifies diverging checkpoint: 
  \STATE $R \rightarrow T_1, T_2$: $d$ s.t. $\tth_j = \tth_j'$ $\forall\ $ $j\leq d \land \tth_{d+1} \neq \tth_{d+1}'$
  \STATE If checkpoint interval length $> 1$: 
  \STATE \textbullet\ Repeat with $\starthash \leftarrow \tth_d$ and $\disphashes \leftarrow [\tth_{d+1}, \tth_{d+1}']$
  \STATE Else: exit loop.
\ENDFOR
\STATE $R$: output $\starthash, \disphashes$ 
as the starting hash and disputed ending hashes of the first diverging training step. 
\end{algorithmic}
\caption{\underline{Dispute Resolution Phase 1:} The \trainers and the referee 
identify the first step the two \trainers diverged at in their executions.}
\label{alg:protocol_phase_1}
\end{algorithm}

{\bf Optimization: multi-level checkpointing.}
We identify the diverging step by narrowing down the interval it occurs in 
through multiple rounds of interaction (see Algorithm~\ref{sec:bisection-phase-1})
During training, the \trainers log checkpoints only at specified steps, instead of every step.
Upon receiving these initial checkpoints sequences, 
the referee identifies the first diverging checkpoint as before.\footnote{Note that we eschew the binary search of Canetti et al., as the number of checkpoint commitments is so low ($N<100$) it is more efficient in practice to send all of them in a single round.}
Then, the parties recurse: they re-run the diverging segment of training and 
log more granular checkpoints within.
These are again sent to the referee 
who identifies and shares the new diverging segment. 
This repeats until the granularity of the checkpoints is a single training step.

\textbf{Communication and storage costs.}
As only short hashes are communicated at this stage (and not the checkpoints themselves), 
the communication overhead is minimal for all parties. 
There is a trade-off between the checkpointing frequency (and thus, the hashing and 
storage requirements) during training and the amount of re-execution needed by the \trainers 
after training during dispute resolution. 
For example, if the trainers log $N$ initial checkpoints 
at each level, 
then the \trainers re-execute a $(\frac{1}{N} + \frac{1}{N^2} + \frac{1}{N^3} + \ldots)$ 
fraction of the original training program. When $N=20$, this 
comes to under $6\%$. 
When considering just the learnable parameters of \llama-8B with precision FP32, 
this requires a few hundred gigabytes of storage.
With $N=100$, the amount of re-execution reduces to under $1.1\%$ but 
the storage requirements reaches a few terabytes. 
As mentioned earlier, the time taken to hash all of the checkpoints is 
negligible compared to the training program itself (and can be done in parallel).

\subsection{Phase 2: Narrowing down the diverging operation.}
\label{sec:bisection-phase-2}

At the end of Phase 1, 
the referee identifies the diverging training step, along with the 
starting commitment of the step (common to \trainers) and the 
two disputed ending commitments. 
To decide which of the two \trainers' claims is dishonest, 
the referee could re-run the entire training step themselves. 
This requires them to receive the starting checkpoint state from either \trainer,
perform the step, 
and compare the hash of the output with that of the two \trainers.
However, this incurs high communication and computation costs for the referee. 
A checkpoint itself can be many gigabytes and 
a training step requires a peak memory usage larger 
than the checkpoint itself. 

To improve on this, we observe that the work done by a training step function is defined as a computational graph.
We can narrow down the dispute further to the 
the first \emph{operator} in the graph 
the two \trainers diverged at. 
Resolving the output of just a single operator can reduce 
the communication and compute requirements of the referee
by two orders of magnitude. 
For example, 
the largest operator involved in most transformer-based models, 
the key-query matrix multiplication in the self-attention layer, 
can be represented as a series of 
matrix-vector multiplication operators, each of which require 
memory in the order of dozens of megabytes even for large sequence lengths. 

\begin{figure}[t]
  \centering
  \includegraphics[width=0.35\textwidth]{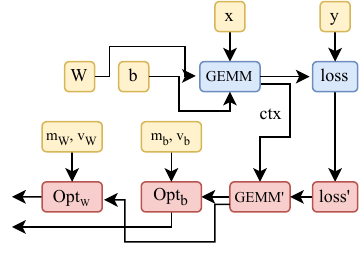}
  \caption{Extended computational graph for a neural network with a single operator. 
Yellow represents nodes that initialize tensor values from either the training data or the 
a training checkpoint. 
    Blue nodes are forward pass operators, and red ones are backward pass operators. 
For clarity, we label the edge transferring context (also called ``saved tensors''' in autograd) 
from the forward pass to the corresponding backward pass operator.
  In the dispute resolution algorithm, these nodes are specified as \cgraphnode objects. 
} 
  \label{fig:extended-graph}
\end{figure}

\textbf{Computational graph and node representations.}
We start with the representation of neural networks as a directed acyclic computational 
graph and extend it to capture all the work done in a training step: 
the forward pass, backward pass, and optimizer state updates.
This extended graph can be implicitly derived 
from the computational graph representing the 
forward pass of the model, such as in a format like \onnx, 
and automatic differentiation library like \autograd.
An example of this expanded graph is shown in Figure \ref{fig:extended-graph}
for a neural network with one operator. 
Note that we topologically sort the graph to ensure a common order for all parties. 

We augment these nodes with information 
allowing the referee to later resolve disputes without needing to 
run other parts of the graph.
On top of information about the node's connections (inputs nodes, output nodes), 
the operation (operator and attribute details), 
we add the hashes of all tensors sent into and emitted out of the node.

\begin{tcolorbox}[colback=azure!5!white, colframe=azure!75!black, title={\cgraphnode}]
\textbullet\ Input node pointers\\
\textbullet\ Output node pointers\\
\textbullet\ Operator and attributes\\
\textbullet\ {\bf Input tensor hashes}: list of hashes of the tensors input to this node \\
\textbullet\ {\bf Output tensor hashes}: list of hashes of the tensors output from this node 
\end{tcolorbox}

If the \trainers start with the same checkpoint 
(which we know to be the case after Phase 1) at the start of a training step 
but finish with different weights or optimizer state, then there must be a node 
in the above computational graph that the \trainers start with the same inputs 
but compute different outputs for. 
Similar to Algorithm~\ref{alg:protocol_phase_1}, we can identify this 
node by comparing the sequence of hashes of the nodes produced by each \trainer. 
When performing a training step during dispute resolution,  
the \trainers populate and hash these nodes, and send the sequence to the 
referee. 
Algorithm~\ref{alg:protocol_phase_2} shows the full protocol.

\begin{algorithm}
  \begin{algorithmic}[1]
\STATE {\bf Participants}: Trainers $T_0, T_1$, Referee $R$
\STATE {\bf Inputs}: checkpoint commitments $\starthash$, $\disphashes$ 
  \STATE $T_1, T_2$ send node hashes to $R$. 
  \STATE \quad\textbullet\ $T_1$ sends ${\tt seq}_0 = \tth_1, \tth_2, \ldots, \tth_k$
  \STATE \quad\textbullet\ $T_2$ sends ${\tt seq}_1 = \tth_1', \tth_2', \ldots, \tth_k$
  \STATE $R$ verifies consistency of the claimed checkpoint states:
  \STATE \quad\textbullet\ $\disphashes[i] \stackrel{?}{=} \text{MerkleHash}({\tt seq}_i)$ for $i=0, 1$
  \STATE $R$ finds and sends the first node index $d$ such that: 
  \STATE \quad\textbullet\ $\tth_j = \tth_j'$ for $j<d$ but $\tth_{d} \neq \tth_{d}'$
  \STATE $T_1, T_2$ open their $d^{th}$ nodes by sending \cgraphnode $\node_0, \node_1$, respectively.  
  \STATE Referee runs decision algorithm using $\node_0, \node_1$. 
  \end{algorithmic}
\caption{\uline{Dispute Resolution Phase 2.} The \trainers and the referee 
identify the first node in the computational graph of the disputed step 
the two \trainers diverged at in their executions.}
\label{alg:protocol_phase_2}
\end{algorithm}

\paragraph{Checkpoint hash format.}

\begin{figure}[h]
  \centering
  \includegraphics[width=0.4\textwidth]{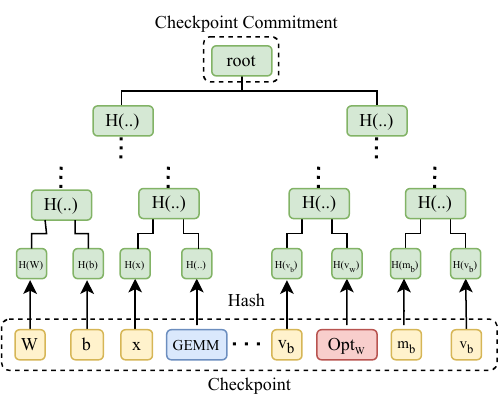}
  \caption{The nodes of the computational graph of the latest training step 
    serve as the checkpoint used in Phase 1. 
    It's committed to using a Merkle (binary hash) tree 
    and verified by the referee in Phase 2 (line {7}).   
    Merkle trees provide efficient proofs of membership for its leaves,
    facilitating efficient dispute resolution when \trainers disagree on the 
    values from the weights, optimizer state, or training data. 
  }
  \label{fig:checkpoint-merkle-tree}
\end{figure}

We can now specify the precise checkpoint format used in Phase 1. 
To ensure continuity between the \trainers claims in Phase 1 and 2, 
the checkpoint after a given training step is the Merkle tree 
with the leaves being the 
\emph{hash of all the {\cgraphnode} nodes} 
of the training step leading to the checkpoint, as shown in Figure~\ref{fig:checkpoint-merkle-tree}.  
As these contain hashes of all updates performed in the step, they serve as a 
commitment to the new training state. 
Importantly, 
they disallow a trainer from using inconsistent commitments between Phase 1 and Phase 2. 
The referee ensures that the sequence of node hashes sent by the 
\trainers in Phase 2 conform exactly to the 
values of $\disphash$ obtained in Phase 1 
(line \textred{7} in Algorithm~\ref{alg:protocol_phase_2}). 

\subsection{Referee's decision algorithm.}
\label{sec:decision}

Finally, we outline the algorithm used by the referee to determine the dishonest party 
in the dispute.
At the end of Phase 2, the referee contains objects $\node_0, \node_1$ representing 
\cgraphnode nodes in the extended computational graph.
As the node hashes are different, there must differ in at least one of the fields 
specified in the \cgraphnode definition. 
As we enforce deterministic execution and control for 
randomness, at most one node can represent a valid execution.
The referee compares the fields within the nodes and determines which of them 
represents a valid execution. 

Depending on how $\node_0$ and $\node_1$ differ, the referee resolves the 
dispute. There are 3 main cases, depending on what differs between the nodes.
\begin{enumerate}
  \item {\bf Case 1:} Difference in graph structure: \texttt{inputs, outputs, operator}  
  \item {\bf Case 2:} Difference in an input tensor hash.
  \item {\bf Case 3:} Difference in an output tensor hash.
\end{enumerate}

We provide brief descriptions of how each case is handled. 
{\bf Case 1} is the easiest 
to resolve as it means one \trainer constructed the graph incorrectly.
The referee, knowing the model specified by the client, can check which structure is correct. 

{\bf Case 2} involves multiple sub-cases 
depending on whether the input tensor corresponding to the differing hash 
is from from (a) the starting checkpoint (such as a weight or 
optimizer state value) or from (b) another node in the graph.
If (a), then the referee requests a \emph{Merkle membership proof} 
of the input hash in the checkpoint from the two \trainers. 
This short and easy-to-check proof can only be produced by the honest \trainer.
If (b), the referee asks for the 
source node's opening from the trainers to check which of the disputed tensor hash 
is present as an output tensor hash of that node.
As nodes $\node_0, \node_1$ are the first diverging node, 
    they agree on source node and in particular, the \texttt{emitted tensor hashes} in it. 
    The disputed input tensor hash must be an emitted tensor hash in the other node. 

{\bf Case 3} implies that the two trainers claim different emitted tensor values. 
This is the only scenario where the referee needs to run the operator 
and decide which \trainer is correct. 
This can be done in multiple ways, depending on the operator and 
compute capabilities of the referee. 
The referee can recompute the operator themselves by receiving from the trainers 
  the input tensors of the node. 
We can also further decompose the operator into smaller operators and perform more phases 
  to narrow down the dispute arbitrarily.

\section{Reproducible Operations (\repops) library}
\label{sec:repops}

Different GPUs, such as those from NVIDIA, AMD, or Intel, 
each come with their own architecture and capabilities.
ML practitioners often tailor to specific hardware, balancing factors like cost, power consumption, 
and computational throughput.
Even between hardware from the same manufacturer, and even when 
adhering to standards like \ieeefp, 
inherent non-determinism in floating-point operations can cause 
numerical discrepancies between different executions of the same program. 
This section describes the source of this non-determinism 
and \repops, 
our library implementing a suite of reproducible ML operators 
which overcome hardware non-determinism.

\subsection{The source of hardware non-determinism}
\label{sec:repops-issue}

Even when following the \ieeefp standard, floating points operations 
are non-associative, meaning the order in which the operations are performed impacts 
the result. 
That is, $(a+b)+c$ can be different from $(a+c)+b$ because each 
addition operation involves rounding that introduces variations. 
These discrepancies arise in parallel programs where functions are split 
and executed 
across multiple threads. 
As GPUs are designed to execute highly parallel code, 
simple differences in architecture, 
such as the number of cores and memory capacity, 
naturally exacerbate this issue.
Even within the same device, storing tensors with 
with different arrangements in memory (that is, the strides)
can alter the sequence of operations, potentially leading to discrepancies.

To overcome this issue, \repops re-implements common ML operators and mathematical 
functions (like exp, sin, cos, tanh) in a way that controls the 
order of 
floating point operators across hardware setups.
On top of this, 
\repops use built-in support for deterministic pseudorandomness generation in \pytorch and \cuda.

\subsection{Controlling the order of operations.} 
\label{sec:repops-order-control}

To enforce a fixed ordering of operations 
while still maximizing parallelism, 
we identify dimensions along which operators can be parallelized 
without introducing non-determinism. 
For dimensions where the order does not affect the outcome, 
parallelization can proceed freely. 
In the dimensions where order is critical, 
we either perform the operations serially or 
synchronize threads to enforce a deterministic execution order. 

We highlight this technique for matrix multiplication, 
which is the most prevalent operator in neural networks. 
Matrix multiplication is represented by three nested for-loops  
where only the third loop introduces potential reordering of floating-point operations.
By parallelizing the first two loops and executing the third loop sequentially 
we can ensure fixed ordering of operations.
\begin{lstlisting}[
]
# repops matrix multiplication
(*@\textcolor{darkspringgreen}{for i = 0 to M-1: \# any order}@*)
    (*@\textcolor{darkspringgreen}{for j = 0 to N-1: \# any order}@*)
        sum = 0
        (*@\textcolor{red}{for k = 0 to K-1: \# fixed order}@*)
            a = A[i][k]
            b = B[k][j]
            sum = sum + a *      b
        (*@\textcolor{black}{C[i][j] = sum}@*)
\end{lstlisting}

Perhaps surprisingly, our benchmarks (Section~\ref{sec:experiments}) reveal that this simple strategy 
incurs manageable overhead.
For matrices larger than  $2^{10}$, our implementation takes 
around 30--60\% extra time as compared 
to non-reproducible matrix multiplication for matrices 
on the GPUs we tested. 
Although this approach may under-utilize parallel processors in some cases, 
it strikes a balance between achieving reproducibility and maintaining performance efficiency.

\subsection{Limitations and future work}

As mentioned, limiting parallelism in the code can under-utilize the compute 
resources available. From our benchmarks 
in Section~\ref{sec:experiments}), one example is matrix multiplication in smaller dimensions,
where the overhead for 256-bit matrices is around $200\%$ on the setups we tested. 
Another is end-to-end training for the relatively smaller \distilbert LLM, 
which incurs a $300\%$ overhead on certain setups.
Mitigating these overheads requires careful tuning and hard-coding of parameters 
for specific hardware.

An inherent assumption for \repops is that 
the hardware adheres to a single floating point operations standard, such as \ieeefp. 
In practice, this limits the hardware and precisions that \repops can run on. 
For example, Nvidia GPUs generally support \ieeefp-compliant FP32 arithmetic, 
whereas that for FP16 is not as widespread. 

\paragraph{Future work: Model parallelism.} 
\repops currently ensures reproducibility between setups when programs are run on a single 
GPU in each setup.
Modern neural network training and inference is 
increasingly done in a distributed setting across multiple GPUs 
using parallelization techniques like data, pipeline, and tensor parallelism. 
Ensuring reproducibility when setups use multiple nodes 
and various types of parallelism 
requires coordinating the division of operators and the order in which 
they're combined during collective communication (such as All-Reduce). 
This can be done using the similar principle of controlled parallelization.
We leave the design and implementation to future work.

\section{Implementation and evaluation}
\label{sec:experiments}
We implement the dispute resolution protocol using the \pytorch framework (version 2.4), with neural network models defined as \onnx computational graphs~\cite{onnx2017}, 
and we implement the \repops library using the \cuda framework (version 12.1). 
\footnote{
A demo of \repops showing reproducibility on certain Llama models 
on CPU and GPU 
is available at \url{https://github.com/gensyn-ai/repop-demo}. 
We're also working on developing \repops into a Python module.
}

This section describe our benchmarks of \repops, focusing on its relative overhead versus 
\pytorch using the hardware-optimized implementations of 
operators from the \cudnn library. 
We test on four GPUs with varying amount of VRAM in our experiments, all from Nvidia: 
T4 (16 GB), RTX 3090 (24 GB), A100 (40 GB variant), and 
A100 (80 GB variant). 
Our \repops implementation currently supports FP32, as that had the most 
widespread \ieeefp compliance support. 

For each experiment, we state our main observations and 
  extract key patterns that shed light on operator performance 
  and indicates directions to take for real-world deployment of \repops. 
We note that the code underlying \cudnn
  is closed-source, and thus we do not have access to all the optimizations 
  they use, nor can we add our own optimizations. 

\subsection{Matrix multiplication.}
\label{sec:matmul-microbenchmarks}
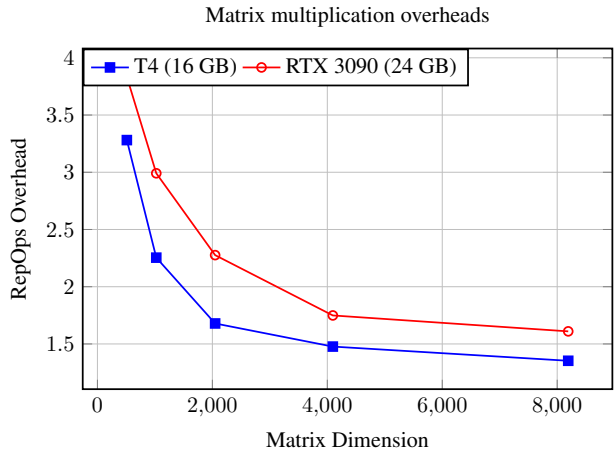
\begin{figure}[t]
    \centering
    \resizebox{\columnwidth}{!}{\pgfplotsset{compat=1.18}

\begin{tikzpicture}
\begin{axis}[
    width=10cm, %
    height=7cm, %
    xlabel={Matrix Dimension}, %
    ylabel={RepOps Overhead}, %
    grid=both, %
    legend style={at={(0,1)}, anchor=north west, legend columns=2},
    title={Matrix multiplication overheads},
    extra y ticks={0},
extra y tick style={
  grid=major,
  major grid style={dashed,line width=0.5pt, black}
}
]

\addplot[
    color=blue, 
    mark=square*, 
    thick
]
coordinates {
    (512, 3.281)
    (1024, 2.254)
    (2048, 1.679)
    (4096, 1.477)
    (8192, 1.353)
};

\addplot[
    color=red, 
    mark=o, 
    thick
] 
coordinates {
    (512, 3.833)
    (1024, 2.991)
    (2048, 2.276)
    (4096, 1.749)
    (8192, 1.610)
};

\legend{T4 (16 GB), RTX 3090 (24 GB)}
\end{axis}
\end{tikzpicture}}
\caption{\repops overhead for matrix-matrix multiplication.}
    \label{fig:matmul_microbenchmarks}
\end{figure}

Matrix multiplication is by far the most prevalent and expensive operation in 
  neural networks, making it a key factor for end-to-end model performance. 
We benchmark the \repops matrix multiplication overhead 
  for various dimensions, 
  comparing it against the \pytorch and \cudnn implementation 
  (i.e., \texttt{torch::mm}). 
For each dimension and hardware pairing, 
we explore a wide array of kernel parameters, 
including elements of the final matrix per thread, tile size, and block sizes, 
and  report the performance of the best one. 
Figure~\ref{fig:matmul_microbenchmarks} shows our results for two GPUs 
emblematic of the general trend we observe. 

\uline{Observation 1: \repops matrix multiplication overhead improves significantly 
with the matrix size.}

The amount of parallelization ``left on the table'' by \repops 
is more significant when the total amount of computation to be done is less, 
as is the case when the matrices are smaller. 
In other words, as matrices grow larger the parallelization performed in \repops 
is enough for better utilization of the GPU's resources. 
This leads to a sort of steady state overhead of 60\%-70\% for RTX 3090 (24 GB) 
and about 35\% for T4 (16 GB).

\subsection{Model inference and training benchmarks.}

We measure the performance of three models, \distilbert (66 million parameters), 
\llama-3.1-1B, and \llama-3.1-8B.
Table~\ref{table:repops_models} shows the average overhead on inference and training tasks 
for the first two models, 
and Table~\ref{table:repops_llama_8b} shows that of inference and fine-tuning tasks for the last
(as our GPUs did not have enough memory to train an 8 billion parameter 
model in FP32). 
We use the Adam optimizer for both training and fine-tuning.
As before, all programs are run in FP32.
We tested on various input shapes and sequence lengths and found the overheads 
obtained to be similar in all settings. 
We tested all training benchmarks on batch sizes from 2 to 8 and
found similar numbers for all benchmarks 
(we show the \emph{worst} performing batch size for each).
We note that the overheads of the \llama models slightly \emph{decreased} as we increase 
the batch size.

\begin{table}[t]
\centering
\caption{\repops training and inference overheads for \distilbert and \llama-1B.}
\label{table:repops_models}
\begin{tabular}{lrrrr}
\toprule
\textbf{Hardware} & \multicolumn{2}{c}{\textbf{\distilbert}} & \multicolumn{2}{c}{\textbf{\llama-1B}} \\
 & \textbf{Infer.} & \textbf{Train.} & \textbf{Infer.} & \textbf{Train.} \\
\midrule
T4 (16 GB)   & 74\% & 258\%  & 218\%  & 374\% \\
A100 (40 GB) & 84\%  & 312\%  & 58\%  & 67\% \\
\end{tabular}
\end{table}

\begin{table}[t]
\centering
\caption{\repops overheads for \llama-8B.}
\begin{tabular}{lcc}
  \toprule 
\textbf{Hardware} & \textbf{Inference} & \textbf{Fine-tuning (LoRA)}   \\
\midrule
A100 (80 GB)   &  98\% & 126\% \\
\end{tabular}
\label{table:repops_llama_8b}
\end{table}

\uline{Observation 2: \repops performs relatively much better on the larger \llama models 
than the smaller \distilbert.}

Analogous to how smaller matrix multiplications incur higher overheads, 
the \distilbert models being less compute intensive accentuates 
the impact of the limited parallelism of \repops. 
Another issue is that \distilbert performs many operators like LayerNorm, GeLU, 
and ERF that are not present in \llama, which we did not focus on optimizing. 

\uline{Observation 3: \llama overheads are significantly better 
in the larger memory (40 GB VRAM) and bandwidth GPU than the smaller memory (16 GB VRAM)
and bandwidth one.}
With memory management being less of a concern, 
larger GPUs see less of the performance issues related to memory transfers as opposed to smaller GPUs. 
Since RepOps as a backend does not tend to allocate excess memory, 
it would make sense that the performance benefits related to larger GPU memory apply not just 
for Torch's runtime but also \repops's. 
This becomes an outsized factor when using larger models with over 1 billion parameters.

\section{Related work}
\label{sec:related-work}

\paragraph{Refereed delegation.} 
Canetti et al.~\cite{canetti13refereed} proposed refereed delegation, 
 with a dispute resolution protocol implementation for CPU (x86) computation.
This idea was adapted to the decentralized setting as a scaling solution for blockchains,
becoming the foundation for ``optimistic rollups'' like TrueBit~\cite{teutsch19scalable}, 
  Arbitrum~\cite{kalodner18arbitrum}, and Optimism~\cite{optimism}.
  
The closest work to our own is Agatha~\cite{zheng18agatha}, which developed a smart contract
to verify neural network inference using a refereed delegation-like protocol.
Our protocol can be viewed as taking this a step further to 
design general dispute resolution protocols 
for both inference and training, which involves capturing a wider variety of dishonest behavior.

\textbf{Hardware non-determinism.} 
To our knowledge, 
  the RepDL library~\cite{repdl_library} is the first open-source implementation of operators made 
  reproducible by controlling the order of floating point operations.
We adopt this approach and expand it to support 
  inference and training for modern LLMs such as \distilbert and the \llama 
  family. 
  
The recent work of \cite{srivastava24optimistic}, also motivated by verifying ML programs,
 tackles the problem of hardware 
non-determinism in a different way, as follows: 
when performing floating point operations in a particular precision (say, in FP32), 
they perform the operator in a higher precision (here, FP64) 
and then round results back to the target precision. 
As rounding can be non-deterministic when 
the numerical error impacts the decision (higher or lower), the party additionally records 
rounding certain rounding decisions they make.
Their assumption is that the final rounded value is deterministic when assuming 
that all numerical discrepancy errors are contained within the higher precision bits.
Thus, when another party re-runs the computation, running all floating point operations 
in the higher precision and 
following the same rounding decisions when specified, they will obtain 
bitwise identical result. 

Our approach with \repops offers several advantages. 
First, the security of Srivastava et al.'s approach assumes that floating point errors 
are always contained in the higher precision bits, which is not guaranteed. 
Additionally \repops works with any lower precision, 
which is crucial as modern machine learning are commonly implemented in lower precision 
(particularly FP16)
to save significant memory and compute resources.
\repops also doesn't require collecting, storing, and transferring the rounding decisions, 
which can be very large: they 18 MB of rounding data per training step 
of the 1.5 billion parameter GPT-2 model.
While \repops incurs a higher training overhead (1.6$\times$ for \llama-1B; 
see Table~\ref{table:repops_models}) 
than their method (1.2-1.4x for GPT-2 as stated in Section 6.1 of ~\cite{srivastava24optimistic}), 
this is mitigated by the ability to use lower precision training.

\section{Concluding discussion}

Our work shows that the refereed delegation model, if sufficiently tailored to the ML setting and with our RepOps library to ensure deterministic computation, provides a middle ground between the expense of fully verified ML training and the insecurity of heuristic approaches.
Bringing our work to practice requires solving several additional practical challenges.
First, to ensure a high likelihood of at least one honest \trainer, a robust ecosystem of \trainers is needed, which are unlikely to collude or suffer related faults (e.g. by running the same third party data center).
Second, incentives are needed to compensate \trainers both for running the original computation and for interacting with the referee to detect dishonest behavior.
The recent BoLD protocol~\cite{alvarez2024bold}, designed for optimistic rollups, may provide a template here.
Finally, our work may see application in the context of blockchains, meaning our referee functionality needs to be implemented to run efficiently as a smart contract in an environment like the Ethereum Virtual Machine (EVM).
EVM was not designed with ML operations involved, and in particular offers no native support for floating point operations.
We leave developing our referee functionality to run in this environment to future work.

\bibliography{refs}

\begin{thebibliography}{16}
\providecommand{\natexlab}[1]{#1}
\providecommand{\url}[1]{\texttt{#1}}
\expandafter\ifx\csname urlstyle\endcsname\relax
  \providecommand{\doi}[1]{doi: #1}\else
  \providecommand{\doi}{doi: \begingroup \urlstyle{rm}\Url}\fi

\bibitem[Abbaszadeh et~al.(2024)Abbaszadeh, Pappas, Katz, and
  Papadopoulos]{abbaszadeh24zero}
Abbaszadeh, K., Pappas, C., Katz, J., and Papadopoulos, D.
\newblock {Zero-Knowledge Proofs of Training for Deep Neural Networks}.
\newblock In \emph{ACM SIGSAC Conference on Computer and Communications
  Security}, CCS '24, New York, NY, USA, 2024. Association for Computing
  Machinery.
\newblock ISBN 9798400706363.
\newblock \doi{10.1145/3658644.3670316}.
\newblock URL \url{https://doi.org/10.1145/3658644.3670316}.

\bibitem[Alvarez et~al.(2024)Alvarez, Arneson, Berger, Bousfield, Buckland,
  Edelman, Felten, Goldman, Jordan, and Kelkar]{alvarez2024bold}
Alvarez, M.~M., Arneson, H., Berger, B., Bousfield, L., Buckland, C., Edelman,
  Y., Felten, E.~W., Goldman, D., Jordan, R., and Kelkar, M.
\newblock {BoLD: Fast and Cheap Dispute Resolution}.
\newblock \emph{arXiv preprint arXiv:2404.10491}, 2024.

\bibitem[Canetti et~al.(2013)Canetti, Riva, and Rothblum]{canetti13refereed}
Canetti, R., Riva, B., and Rothblum, G.~N.
\newblock Refereed delegation of computation.
\newblock \emph{Information and Computation}, 226:\penalty0 16--36, 2013.
\newblock ISSN 0890-5401.
\newblock \doi{https://doi.org/10.1016/j.ic.2013.03.003}.
\newblock URL
  \url{https://www.sciencedirect.com/science/article/pii/S0890540113000217}.
\newblock Special Issue: Information Security as a Resource.

\bibitem[Chetlur et~al.(2014)Chetlur, Woolley, Vandermersch, Cohen, Tran,
  Catanzaro, and Shelhamer]{chetlur14cuDNN}
Chetlur, S., Woolley, C., Vandermersch, P., Cohen, J.~M., Tran, J., Catanzaro,
  B., and Shelhamer, E.
\newblock {cuDNN: Efficient Primitives for Deep Learning}.
\newblock \emph{ArXiv}, abs/1410.0759, 2014.
\newblock URL \url{https://api.semanticscholar.org/CorpusID:12330432}.

\bibitem[Choi et~al.(2024)Choi, Shavit, and Duvenaud]{choi24tools}
Choi, D., Shavit, Y., and Duvenaud, D.
\newblock Tools for verifying neural models' training data.
\newblock In \emph{37th International Conference on Neural Information
  Processing Systems}, NeurIPS '23, 2024.

\bibitem[Fang et~al.(2023)Fang, Jia, Thudi, Yaghini, Choquette-Choo, Dullerud,
  Chandrasekaran, and Papernot]{fang23proof}
Fang, C., Jia, H., Thudi, A., Yaghini, M., Choquette-Choo, C.~A., Dullerud, N.,
  Chandrasekaran, V., and Papernot, N.
\newblock {Proof-of-Learning is Currently More Broken Than You Think}.
\newblock In \emph{8th IEEE European Symposium on Security and Privacy
  (EuroS\&P)}. IEEE, 2023.
\newblock ISBN 978-1-6654-6512-0.
\newblock \doi{10.1109/EuroSP57164.2023.00052}.
\newblock URL \url{https://doi.org/10.1109/EuroSP57164.2023.00052}.

\bibitem[Jia et~al.(2021)Jia, Yaghini, Choquette-Choo, Dullerud, Thudi,
  Chandrasekaran, and Papernot]{jia21proofoflearning}
Jia, H., Yaghini, M., Choquette-Choo, C.~A., Dullerud, N., Thudi, A.,
  Chandrasekaran, V., and Papernot, N.
\newblock {Proof-of-Learning: Definitions and Practice}.
\newblock \emph{IEEE Symposium on Security and Privacy (S\&P)}, pp.\
  1039--1056, 2021.
\newblock URL \url{https://api.semanticscholar.org/CorpusID:232168663}.

\bibitem[Kalodner et~al.(2018)Kalodner, Goldfeder, Chen, Weinberg, and
  Felten]{kalodner18arbitrum}
Kalodner, H.~A., Goldfeder, S., Chen, X., Weinberg, S.~M., and Felten, E.~W.
\newblock Arbitrum: Scalable, private smart contracts.
\newblock In \emph{USENIX Security Symposium}, 2018.
\newblock URL \url{https://api.semanticscholar.org/CorpusID:52046120}.

\bibitem[Kingma \& Ba(2014)Kingma and Ba]{kingma14adam}
Kingma, D.~P. and Ba, J.
\newblock Adam: A method for stochastic optimization.
\newblock \emph{arXiv preprint arXiv:1412.6980}, 2014.

\bibitem[Microsoft(2023)]{repdl_library}
Microsoft.
\newblock Repdl.
\newblock \url{https://github.com/microsoft/RepDL/tree/main}, 2023.
\newblock Accessed: 2025-01-06.

\bibitem[{ONNX}(2017)]{onnx2017}
{ONNX}.
\newblock Onnx: Open neural network exchange, 2017.
\newblock URL \url{https://onnx.ai}.
\newblock Accessed: 2025-01-30.

\bibitem[{Optimism}(2025)]{optimism}
{Optimism}.
\newblock Rollup protocol overview.
\newblock \url{https://docs.optimism.io/stack/rollup/overview}, 2025.
\newblock Accessed on January 29, 2025.

\bibitem[Srivastava et~al.(2024)Srivastava, Arora, and
  Boneh]{srivastava24optimistic}
Srivastava, M., Arora, S., and Boneh, D.
\newblock {Optimistic Verifiable Training by Controlling Hardware
  Nondeterminism}.
\newblock \emph{ArXiv}, abs/2403.09603, 2024.
\newblock URL \url{https://api.semanticscholar.org/CorpusID:268385340}.

\bibitem[Sun et~al.(2024)Sun, Li, and Zhang]{sun24zkllm}
Sun, H., Li, J., and Zhang, H.
\newblock {zkLLM: Zero Knowledge Proofs for Large Language Models}.
\newblock In \emph{ACM SIGSAC Conference on Computer and Communications
  Security}, CCS '24, pp.\  4405–4419, 2024.
\newblock ISBN 9798400706363.
\newblock \doi{10.1145/3658644.3670334}.
\newblock URL \url{https://doi.org/10.1145/3658644.3670334}.

\bibitem[Teutsch \& Reitwie{\ss}ner(2019)Teutsch and
  Reitwie{\ss}ner]{teutsch19scalable}
Teutsch, J. and Reitwie{\ss}ner, C.
\newblock A scalable verification solution for blockchains.
\newblock \emph{CoRR}, abs/1908.04756, 2019.
\newblock URL \url{http://arxiv.org/abs/1908.04756}.

\bibitem[Zheng et~al.(2021)Zheng, Xie, Zhang, Chen, Chen, Guo, Sun, Sun, and
  Zhou]{zheng18agatha}
Zheng, Z., Xie, P., Zhang, X., Chen, S., Chen, Y., Guo, X., Sun, G., Sun, G.,
  and Zhou, L.
\newblock Agatha: Smart contract for {DNN} computation.
\newblock \emph{CoRR}, abs/2105.04919, 2021.
\newblock URL \url{https://arxiv.org/abs/2105.04919}.

\end{thebibliography}
\bibliographystyle{icml2025}

\newpage
\appendix
\onecolumn

\end{document}